# Guided Evolutionary Neural Architecture Search With Efficient Performance Estimation

Vasco Lopes[a,c,*], Miguel Santos[a], Bruno Degardin[b,c] and Luís A. Alexandre[a]

[a]*NOVA Lincs, Universidade da Beira Interior, Covilhã, Portugal*
[b]*Instituto de Telecomunicações, Covilhã, Portugal*
[c]*DeepNeuronic, Covilhã, Portugal*

ABSTRACT

Neural Architecture Search (NAS) methods have been successfully applied to image tasks with excellent results. However, NAS methods are often complex and tend to converge to local minima as soon as generated architectures seem to yield good results. This paper proposes GEA, a novel approach for guided NAS. GEA guides the evolution by exploring the search space by generating and evaluating several architectures in each generation at initialisation stage using a zero-proxy estimator, where only the highest-scoring architecture is trained and kept for the next generation. Subsequently, GEA continuously extracts knowledge about the search space without increased complexity by generating several off-springs from an existing architecture at each generation. More, GEA forces exploitation of the most performant architectures by descendant generation while simultaneously driving exploration through parent mutation and favouring younger architectures to the detriment of older ones. Experimental results demonstrate the effectiveness of the proposed method, and extensive ablation studies evaluate the importance of different parameters. Results show that GEA achieves state-of-the-art results on all data sets of NAS-Bench-101, NAS-Bench-201 and TransNAS-Bench-101 benchmarks.

## 1. Introduction

Convolutional Neural Networks (CNNs) have been extensively applied with success to a panoply of tasks with unprecedented results, from image classification [1, 2], to semantic segmentation [3], text analysis [4], amongst many others [5]. Their inherent feature extraction capabilities allow CNNs to be easily applied and transferred to different problems. Over the years, several carefully designed architectures have incrementally out-performed the state-of-the-art by proposing novel components and mechanisms, such as skip and residual connections, faster and less size intensive operations and attention mechanisms [6, 7, 8, 9, 10, 11, 12]. However, designing tailor-made highly performant CNNs for a given problem is a grueling endeavour. The design choices intrinsic to the architectures, layer combinations and training require extensive architecture engineering, which is heavily dependant on human expertize and trial and error. Thus, a logical step was to start automating the architecture engineering and design, creating a growing interest in Neural Architecture Search (NAS).

NAS has been successfully applied in designing architectures for image and text problems [13, 14, 15, 16, 17, 18, 19]. Commonly, NAS proposals are composed of three components. First, the search space, which specifies the possible operations to be sampled and their connections, ultimately defining the type of architectures that the search method can generate. Secondly, the search method, which represents the approach used to explore the search space and

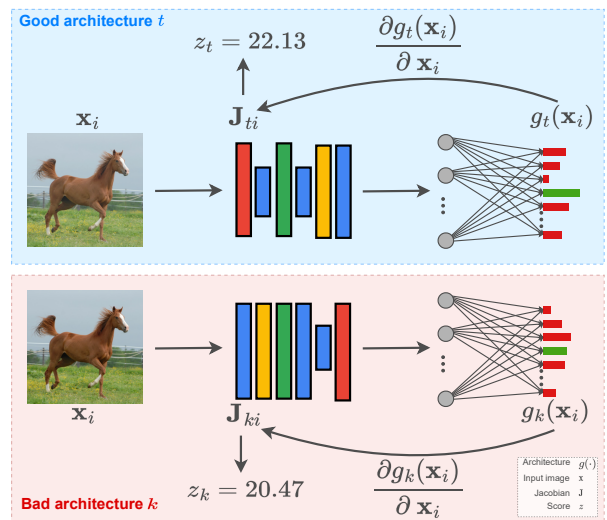

**Figure 1:** GEA example of scoring two different architectures using the same input. Generated architectures at each generation are ranked based on a score that correlates with their final performance, which determines which architecture is selected to be part of the population.

generate architectures. The most common approaches are reinforcement learning, evolutionary strategies and gradient-based methods, which commonly work by updating a controller to sample more efficient architectures based on the performance of the generated models. Finally, the performance estimation strategy, which defines how the generated architectures are evaluated. Thus, the goal of a NAS method is to, based on the search method, efficiently search a large

[*]Corresponding author
✉ vasco.lopes@ubi.pt (V. Lopes)
ORCID(s): 0000-0002-5577-1094 (V. Lopes); 0000-0003-2462-7310 (B. Degardin); 0000-0002-5133-5025 (L.A. Alexandre)

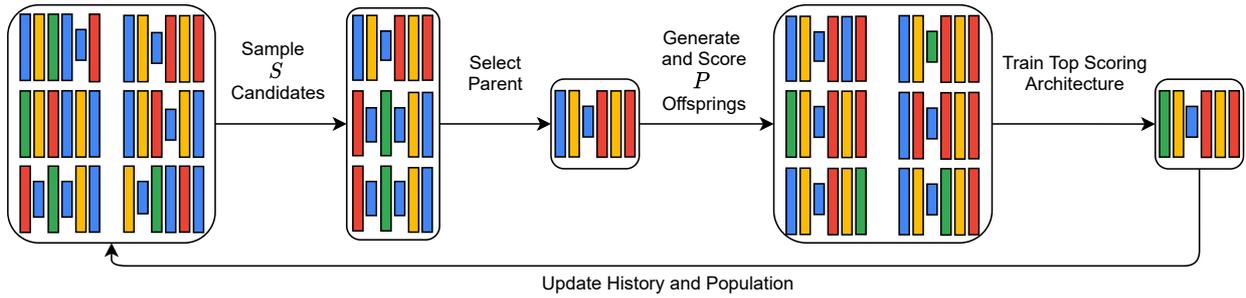

**Figure 2:** Illustration of one iteration of GEA. Architectures are represented with varying bar's length and colors to illustrate their diversity. The process shows the sampling of several candidates, parent selection, offspring generation and evaluation and top scoring architecture training.

set of possible networks to find an optimal architecture for a given problem.

Despite excellent results obtained by prominent NAS methods, the computational cost of most approaches is high, which in some cases can be in the order of months of GPU computation [20, 21, 22]. To mitigate this, approaches focus on a cell-based design, where NAS methods design small cells that are replicated through an outer-skeleton, thus alleviating the complexity of the search space [23, 20, 22, 24, 25]. Furthermore, several performance estimation strategies have been proposed to reduce the time constraint of NAS methods, by mainly conducting low-fidelity estimates, learning curve extrapolations, statistical approaches [26, 13] or by proposing one-shot methods, where the weights of the generated models are inherited from a super-network [27, 28, 29]. However, searching through high-dimensional search spaces is highly complex, even when there is some prior knowledge about the space. Most prominent NAS methods fail to generalise to new data sets due to fast convergence to local minima [30, 31], thus jeopardizing the search and method's applicability. The most reliable approach to obtain information about the search space while searching is to fully train generated architectures and optimise the search based on the most performant ones. However, this is extremely costly, and results are highly dependant on the training schemes and initialisation setups [31]. Therefore, zero-proxy estimators present an attractive solution, where statistics are drawn from the generated architectures to score them at initialisation stage, thus requiring no training [32, 33]. These methods are time efficient and capable of performing good correlations between the score and respective accuracies when the architectures are trained [34, 26].

This paper proposes GEA, an evolutionary NAS method that leverages zero-proxy estimation to efficiently guide the search. By using an evolutionary strategy where operations can be mutated and younger architectures are prefered, GEA forces an exploitation of the most performant architectures, and an exploration of the search space by performing mutations. More, we solve the problem of conducting full evaluation of the generated architectures to obtain knowledge about the search space by generating several architectures in each generation, where all are evaluated at initialisation stage using a zero-proxy estimator and only the highest scoring architecture is trained and kept for the next generation. By doing so, GEA is capable of continuously extracting knowledge about the search space without compromising the search, resulting in state-of-the-art results in NAS-Bench-101, NAS-Bench-201 and TransNAS-Bench-101 search spaces. Figure 1 shows the process of evaluating generated architectures. The code is publicly available at https://github.com/VascoLopes/GEA.

Our contributions can be summarized as:

- We propose a guided NAS method based on evolutionary strategies and zero-proxy estimation to generate image classifier architectures - Convolutional Neural Networks.

- We empirically show that guided mechanisms can be used without compromising time efficiency nor the generated models performance. Also, we detail the algorithm, emphasizing the accessible transferability of the guiding mechanism.

- We achieve state-of-the-art results on all data sets of NAS-Bench-101, NAS-Bench-201 and TransNAS-Bench-101 benchmarks, thus showing GEA's generability.

- Extensive ablation studies show the importance of different parameters and regularization, thus shedding insights for the design of NAS evolutionary models.

## 2. Related Work

NAS was initially proposed as a Reinforcement Learning (RL) problem, where a controller is trained based on the generated architecture's performances to incrementally sample more efficient ones [20]. Follow-up approaches focused on improving the overall performance, and the computation required to frame NAS as a RL problem by proposing the use of different learning strategies, distributed computing, and novel incremental sampling strategies [35, 13, 14, 36]. ENAS [28], showed that RL could be used to perform NAS in a reasonable time-frame by training a controller to discover architectures through optimal subgraph search within a large computational graph, requiring only a few computational days. DARTS [27], proposed the use of gradients to generate architectures by performing continuous relaxation of the parameters using a bi-level gradient optimization, resulting in the generation of competitive architectures in a few GPU days. These methods served as basis for follow-up weight-sharing NAS methods and one-shot models [24, 37, 38, 39, 40, 29].

Evolutionary computation is a common approach for NAS. NEAT was one of the first evolutionary methods to evolve simple neural networks [41], which served as base and inspiration for methods that evolve deeper architectures where parent architectures have their parameters mutated to force evolution towards better performances [42, 43]. REA was one of the first evolutionary NAS methods to be proposed [44]. It evolves architectures through mutations, and employs a tournament selection to serve as a regularization mechanism for the population. Although several evolutionary algorithms have been proposed to incrementally improve the performance of the method by designing novel heuristics to perform the evolution [45, 46, 47], most proposals are still computationally heavy, requiring several days or weeks of computation, and quickly converge to non-optimal minima.

Guiding mechanisms have been proposed to improve NAS. PNAS introduced consortium learning to improve the search, where the design of architectures is gradual, based on the evaluation of increasingly larger networks [21]. This approach allowed the method to be progressively guided through the search space, training only a portion of the architectures based on the estimation of the performance by a predictor network. However, this method still required immense computation. NPENAS guides an evolutionary search by proposing two predictors: a graph-based uncertainty estimation network and a performance predictor [48]. In [49], the authors evaluate the similarity of the internal activations of the generated architectures against a known one, e.g., ResNet, via representational similarity analysis to extract information about the search. [50] proposes the use of landmark architecture evaluation to regularize the ranking of child architectures in super-net settings, thus guiding the search towards a better ranking correlation between stand-alone architectures and the super-net ranking.

In this work, we leverage the findings that show that evolving architectures are an efficient approach for NAS [44], and that zero-proxy estimators provide a reasonably good and extremely fast scoring of untrained architectures [32]. By coupling a zero-proxy estimator as a guiding mechanism to the search method, we force further exploitation of settings favourable to the generated architectures. At the same time, it also efficiently allows the exploration of the search space by quickly evaluating thousands of architectures, thus providing vital information to guide the search.

## 3. Proposed Method

The goal of a NAS algorithm is to find an optimal architecture $a^*$ from the space of architectures $\mathcal{A}$, $a^* \in \mathcal{A}$, such

---

**Algorithm 1** Guided Evolution

$population \leftarrow$ empty queue      ▷ Population.
$history \leftarrow \emptyset$      ▷ Models history.
**while** #$population < C$ **do**      ▷ Initialize population.
     $model.arch \leftarrow$ RANDOMARCHITECTURE()
     $model.accuracy \leftarrow$ ZEROPROXY($model.arch$)
     add $model$ to right of $population$    ▷ Force model age
**end while**
drop the $C - P$ worst individuals from $population$
**for** $model \in population$ **do**
     $model.accuracy \leftarrow$ TRAINANDEVAL($model.arch$)
     add $model$ to history
**end for**
**while** #$history < C$ **do**
     $sample \leftarrow S$ random candidates from the population (with replacement)
     $parent \leftarrow$ highest-accuracy model in $sample$
     $generation \leftarrow \emptyset$
     **while** #$generation < P$ **do**
         $child.arch \leftarrow$ MUTATE($parent.arch$)
         $child.accuracy \leftarrow$ ZEROPROXY($child.arch$)
         add $child$ to $generation$
     **end while**
     $top\_child \leftarrow$ highest-performant model in generation
     $top\_child.accuracy \leftarrow$ TRAINANDEVAL($top\_child.arch$)
     add $top\_child$ to right of $population$
     add $top\_child$ to $history$
     remove $dead$ from left of $population$    ▷ Oldest model.
**end while**
**return** highest-accuracy model in $history$    ▷ Most performant.

---

that it maximizes an objective function $\mathcal{O}$. The proposed method, GEA, frames NAS as an optimization problem where an evolutionary strategy evolves architectures $a \in \mathcal{A}$ based on operation mutations and guided evolution. Therefore, we can define our problem as a nested optimization problem, where the goal is to find a final network, $\mathbb{L}$, created from training the optimal architecture, $a^*$, on a training set, $d^{(train)}$, such that $a^*$ maximizes an objective function $\mathcal{O}$ for a given task, on the validation set:

$$a^* = \arg\max_{a \in \mathcal{A}} \mathcal{O}(\mathbb{L}(a, d^{(train)}), d^{(valid)}) \qquad (1)$$

In the following sections, we detail GEA and the zero-proxy estimator leveraged to create the guiding mechanism.

### 3.1. Search Method

GEA is summarised in Algorithm 1. In detail, GEA starts by randomly generating $C$ architectures from the search space of possible architectures, $\mathcal{A}$. The architectures that belong to the search space have equal probabilities of being randomly sampled. Sampled architectures are then evaluated using a zero-proxy estimator that scores the architectures at initialisation stage, without requiring any training (the zero-proxy estimation mechanism is detailed in section 3.2). Then, from the $C$ scored architectures, only the top $P$ scoring ones are added to the population and trained to extract their fitness, $f$. The fitness, $f$, is the validation accuracy after a partial train (few epochs). By scoring $C$ architectures at initialisation stage, GEA acquires knowledge regarding the search space, which is then exploited by selecting the top performant architecture, thus guiding the upcoming search by weeding out bad architectures.

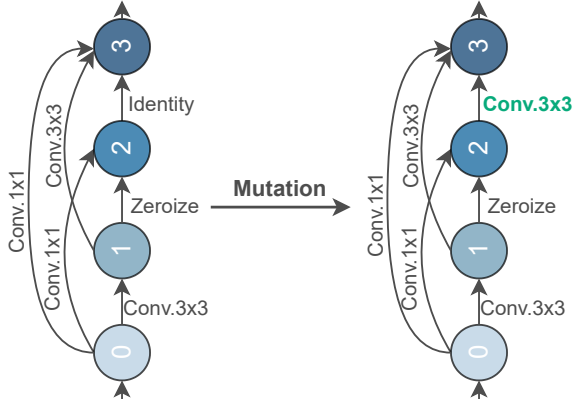

**Figure 3:** Example of mutating one operation of a cell using the NAS-Bench-201 search space: operation Identity becomes Conv. $3 \times 3$.

Once the initial population is defined, the evolution takes place for $C$ cycles. At each iteration, the first step is to perform a tournament selection. For this, $S$ architectures are randomly and uniformly sampled from the population. Then, the architecture with the highest fitness score, $f$, from the pool of $S$ architectures is selected to be the parent of the next generation (cycle). To generate new architectures, GEA performs a mutation over the parent architecture. The mutation works by randomly changing one operation of the architecture by another from the pool of operations. An example of a mutation using the NAS-Bench-201 search space is visually represented in Figure 3. $P$ new architectures are generated at each cycle by performing operation mutations over the selected parent, which are then scored using the zero-proxy estimator. The highest-scoring architecture is kept and added to the population after evaluating its fitness. Generating and evaluating $P$ architectures strengthens the search method to find the best direction for the parent's evolution across the search space. This allows the method to be guided through a complex space without jeopardizing the time required to perform the evolution or the search method's complexity. When the new architecture is added to the population, a regularization mechanism (survivor selection) takes place, where the oldest architecture is removed and discarded, thus forcing exploration of the search space by favouring younger architectures that represent new settings evolved by prior acquired knowledge.

Inherently, higher $P$ values represent a higher degree of exploration of the search space, while higher $S$ values represent higher exploitation by increasing the probability of the best architectures in the population being selected as parents for the next generation.

### 3.2. Zero-proxy Estimator

The goal of scoring architectures at initialisation stage is to provide information about the search space without incurring in the high cost of actually training them, thus allowing guiding the search to optimal settings. For this, we use a zero-proxy estimator based on Jacobian covariance. This allows us to quickly evaluate if an architecture is good without requiring any training, thus allowing the selection of a generated architecture to be added to the population with more confidence that the search is being correctly guided. To do this, we can define a linear mapping, $w_i = g(\mathbf{x}_i)$, which maps the input $\mathbf{x}_i \in \mathbb{R}^D$, through the network, $g(\mathbf{x}_i)$, where $\mathbf{x}_i$ represents an image that belongs to a batch $\mathbf{X}$, and $D$ is the input dimension [32]. Then, the Jacobian of the linear map can be computed using:

$$\mathbf{J}_i = \frac{\partial g(\mathbf{x}_i)}{\partial \mathbf{x}_i} \qquad (2)$$

This allows us to evaluate the architecture's behaviour for different images by calculating $\mathbf{J}_i$ for different data points, $g(\mathbf{x}_i)$, of a single batch $\mathbf{X}$, $i \in 1, \cdots, N$:

$$\mathbf{J} = \left( \frac{\partial g(\mathbf{x}_1)}{\partial \mathbf{x}_1} \quad \frac{\partial g(\mathbf{x}_2)}{\partial \mathbf{x}_2} \quad \cdots \quad \frac{\partial g(\mathbf{x}_N)}{\partial \mathbf{x}_N} \right)^\top \qquad (3)$$

$\mathbf{J}$ contains information about the architecture's output with respect to the input for several images. We can split this into classes and evaluate how an architecture models complex functions at initialisation stage and its effect on images that belong to the same class. To do that, we split $\mathbf{J}$ into several sets, where each set, $\mathbf{M}_k$, contains all $\mathbf{J}_i$ that belong to the same class $k$. Then, we can calculate a per-class correlation matrix, $\mathbf{\Sigma}_{\mathbf{M}_k}$, using the obtained sets, $\mathbf{M}_k$, where $k = 1, ... K$.

Individual correlation matrices provide information about how a single architecture treats images for each class. However, different correlation matrices might yield different sizes, as the number of images per class differ. To be able to compare different correlation matrices, they are individually evaluated:

$$\mathbf{E}_k = \begin{cases} \sum_{i=1}^{N} \sum_{j=1}^{N} log(|(\mathbf{\Sigma}_{\mathbf{M}_k})_{i,j}| + t), & \text{if } K \leq \tau \\ \dfrac{\sum_{i=1}^{N} \sum_{j=1}^{N} log(|(\mathbf{\Sigma}_{\mathbf{M}_c})_{i,j}| + t)}{\sqrt{\#\mathbf{\Sigma}_{\mathbf{M}_k}}}, & \text{otherwise} \end{cases} \qquad (4)$$

where $t$ is a small-constant with the value of $1 \times 10^{-5}$, and $K$ is the number of classes in batch $\mathbf{X}$, and # represents the number of elements.

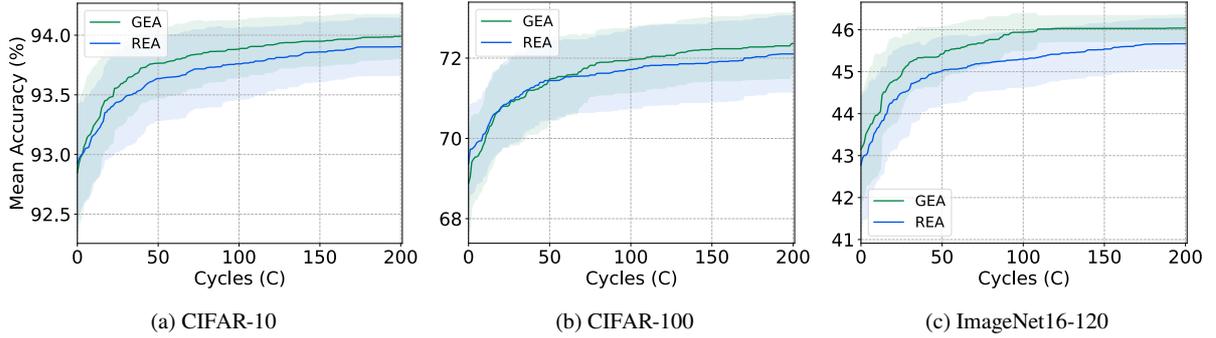

**Figure 4:** Mean accuracy and standard deviation over 25 runs of the proposed method, GEA, and direct comparison with REA for different cycles ($C$) across CIFAR-10, CIFAR-100 and ImageNet16-120 data sets.

**Table 1**
Mean test accuracy (%) and standard deviation across 50 runs in NAS-Bench-101 CIFAR-10 data set. Evaluation for REA and GEA was done with P/S/C=10/5/200.

| Method | Search Time (s) ↓ | Mean Test Accuracy (%) ↑ |
|---|---|---|
| RS | N/A | 90.38±5.51 |
| REA | 26676.49 | 93.12±0.48 |
| **GEA (ours)** | 30128.32 | **93.99±0.25** |

Finally, an architecture is scored based on the individual evaluations of the correlation matrices by:

$$z = \begin{cases} \sum_{w=1}^{K} |\mathbf{e}_w|, & \text{if } K \leq \tau \\ \frac{\sum_{i=1}^{K} \sum_{j=i+1}^{K} |\mathbf{e}_i - \mathbf{e}_j|}{\#\mathbf{e}}, & \text{otherwise} \end{cases} \quad (5)$$

where **e** contains all the correlation matrices' scores. The final score is dependant on the number of classes present in **X**, as data sets with a higher number of classes commonly have more noise, which is mitigated by conducting a normalized pair-wise difference. We empirically defined $\tau = 100$, based on the search space and data sets used.

We can then use $z$ to rank the generated architectures, providing an efficient mechanism of differentiating between bad and good architectures, thus allowing the search to be guided towards better settings without compromising the search cost.

## 4. Experiments
### 4.1. Search Spaces

To evaluate the effectiveness of the proposed NAS algorithm, we utilise three different search spaces: NAS-Bench-101 [51], NAS-Bench-201 [30] and TransNAS-Bench-101 [52] benchmarks. These benchmarks were designed to have tractable NAS search spaces with metadata for the training of thousands of architectures within those search spaces.

**NAS-Bench-101** is a cell-based search space consisting of 423,624 neural networks that have been trained, with three different initialisations, on CIFAR-10 for 108 epochs each. In NAS-Bench-101 search space there are three possible operations: $1 \times 1$ and $3 \times 3$ convolution and $3 \times 3$ max pooling. Convolution operations are combined with batch normalization and ReLU operations to create a Conv-BN-ReLU pattern. To form entire architectures, each cell is initially stacked 3 times, followed by a max-pooling layer that serves the purpose of halving the image height and width, and doubling the number of channels. This pattern is repeated 3 times and followed by a global average pooling and a final classification layer with a softmax function.

**NAS-Bench-201** fixes the search space as a cell-based design with 5 possible operations: zeroize, skip connection, $1 \times 1$ and $3 \times 3$ convolution, and $3 \times 3$ average pooling layer. The cell design comprises six edges and four nodes, where an edge represents a possible operation through two nodes. By fixing the cell size and the operation pool, the search space comprises $5^6 = 15625$ possible cells. To form entire networks, the cells are replicated in an outer-defined skeleton. NAS-Bench-201 provides information regarding the training and performance of all possible networks in the search space in three different data sets: CIFAR-10, CIFAR-100 and ImageNet16-120, thus proposing a controlled setting that allows different NAS methods to be fairly compared, as they are forced to use the search space, training procedures and hyper-parameters.

**TransNAS-Bench-101** is a benchmark that provides architecture's performances across seven vision tasks including classification, regression, pixel-level prediction and self-supervised tasks. The 7 tasks of this benchmark are: object classification, scene classification, autoencoding, surface normal, semantic segmentation, room layout and jigsaw. By having multiple tasks that are queryable with the same input, this benchmark provides the opportunity to evaluate NAS transferability between different tasks. There are two types of search space in this benchmark, i.e., the widely-studied cell-based search space containing 4096 architectures and macro skeleton search space based on residual blocks containing 3256 architectures. Possible operations are: zeroize, skip connection, 1×1 and 3×3 convolution. Transnas-Bench-101 provides information regarding the training and performance of all possible networks in the search space using the same training protocols and hyper-parameters within each task.

**Table 2**
Comparison of manually designed networks and several search methods evaluated using the NAS-Bench-201 benchmark. Performance is shown in terms of accuracy (%) with mean±std, on CIFAR-10, CIFAR-100 and ImageNet-16-120. Search times are the mean time required to search for cells in CIFAR-10. Search time includes the time taken to train networks as part of the process where applicable. Table adapted from [30, 32, 33].

| Method | Search Time (s)↓ | CIFAR-10 | | CIFAR-100 | | ImageNet-16-120 | |
|---|---|---|---|---|---|---|---|
| | | Val. Acc (%)↑ | Test Acc. (%)↑ | Val. Acc (%)↑ | Test Acc. (%)↑ | Val. Acc (%)↑ | Test Acc. (%)↑ |
| **Manually designed** | | | | | | | |
| ResNet | - | 90.83 | 93.97 | 70.42 | 70.86 | 44.53 | 43.63 |
| **Weight sharing** | | | | | | | |
| RSPS | 7587 | 84.16±1.69 | 87.66±1.69 | 59.00±4.60 | 58.33±4.34 | 31.56±3.28 | 31.14±3.88 |
| DARTS-V1 | 10890 | 39.77±0.00 | 54.30±0.00 | 15.03±0.00 | 15.61±0.00 | 16.43±0.00 | 16.32±0.00 |
| DARTS-V2 | 29902 | 39.77±0.00 | 54.30±0.00 | 15.03±0.00 | 15.61±0.00 | 16.43±0.00 | 16.32±0.00 |
| GDAS | 28926 | 90.00±0.21 | 93.51±0.13 | 71.14±0.27 | 70.61±0.26 | 41.70±1.26 | 41.84±0.90 |
| SETN | 31010 | 82.25±5.17 | 86.19±4.63 | 56.86±7.59 | 56.87±7.77 | 32.54±3.63 | 31.90±4.07 |
| ENAS | 13315 | 39.77±0.00 | 54.30±0.00 | 15.03±0.00 | 15.61±0.00 | 16.43±0.00 | 16.32±0.00 |
| **Non-weight sharing** | | | | | | | |
| RS | 12000 | 90.93±0.36 | 93.70±0.36 | 70.93±1.09 | 71.04±1.07 | 44.45±1.10 | 44.57±1.25 |
| REINFORCE | 12000 | 91.09±0.37 | 93.85±0.37 | 71.61±1.12 | 71.71±1.09 | 45.05±1.02 | 45.24±1.18 |
| BOHB | 12000 | 90.82±0.53 | 93.61±0.52 | 70.74±1.29 | 70.85±1.28 | 44.26±1.36 | 44.42±1.49 |
| REA† | 26070 | 91.22±0.25 | 93.97±0.31 | 72.36±1.07 | 72.14±0.86 | 45.09±0.92 | 45.55±1.02 |
| **GEA (ours)**† | 26911 | **91.26±0.20** | **93.99±0.23** | **72.62±0.77** | **72.36±0.66** | **45.97±0.72** | **46.04±0.67** |

† Results of 25 runs using the same settings: $P/S/C = 10/5/200$, using a single 1080Ti GPU.

### 4.2. Results and Discussion

First, we evaluate the proposed method on NAS-Bench-101. For this, we fixed $P/S/C = 10/5/200$, following standard settings used and assessed by prior works [30, 44], and directly compare it against random search and REA [44]. In Table 1 we present this comparison in terms of search cost, in seconds, and mean test accuracy and standard deviation, calculated from executing GEA and REA 50 times. From the results, it is clear that GEA outperforms REA and heavily improves when compared to RS. GEA is highly efficient, requiring only approximately 0.3 GPU days to execute each run. The results show that the guiding mechanism can improve the search, promoting regions that yield better architectures in terms of accuracy.

Then, we evaluate GEA using the NAS-Bench-201 search space. The first experiment in this search space was to directly compare GEA with REA for a different number of generations/cycles, $C$. This also allows the evaluation of the importance of $C$, which is the main parameter that inherently defines the time required for the search procedure. Higher $C$ values will take longer to finish. More, $C$ establishes the number of architectures that are evaluated: $C \times P$ architectures ($P$ per cycle) are generated and evaluated using the zero-proxy estimation method to provide information about the search space, from which $C$ architectures (1 per cycle) are selected and trained. The results are expressed as the mean test accuracy and standard deviation as colored areas, obtained by the best architecture found by each method for different $C$ values over 25 different runs, and are presented in Figure 4. In this experiment, the $P/S$ used to allow a fair comparison was set to $P/S = 10/5$, following the typical settings used by prior works [30, 44]. The results show that across all data sets, GEA consistently outperforms REA, and is capable of converging to better results even for small numbers of $C$. These results demonstrate that the search method converges more quickly to regions of the search space that contain better architectures by leveraging the guided mechanism. Also, on ImageNet16-120, the noisier data set on NAS-Bench-201, the result from the T-test analysis was $\rho = 0.033$, thus showing statistical significance between the results obtained by GEA when compared to REA. Note that for our proposed method, GEA, $P$ value means that at any given time of the search, the population is equal to 10 architectures and that from the pool of parents, 5 architectures are sampled with replacement to select the parent of the generated architectures at a given cycle. The sampled parent generates $P$ architectures through mutation per cycle, which are evaluated using the zero-proxy estimator, wherein only the top-scoring architecture is selected to integrate the population. By selecting $S > 1$ architectures to have the opportunity of being a parent, we are leveraging the intrinsic exploitation characteristics of the evolutionary strategy, whereas by generating $P$ architectures, we are forcing exploitation that guides the search more effectively.

In Table 2 we further compare GEA, using $P/S/C = 10/5/200$, against other state-of-the-art methods on the NAS-Bench-201 search space, using as evaluation metrics the mean accuracy, standard deviation, and search time in seconds, across the 3 data sets. GEA consistently outperforms both weight sharing and non-weight sharing NAS methods, achieving state-of-the-art results in all three data sets. Moreover, GEA is extremely efficient in terms of search time, requiring only 0.3 GPU days to complete the search. Even though GEA evaluates $C \times P$ architectures with the zero-proxy estimator and further evaluates $C$ architectures by partially training them, it requires a similar search time as REA under the same settings and considerably less than most weight sharing methods. Lower standard deviation also indicates that GEA is precise and capable of generating high performant architectures, which is especially valid in ImageNet16-120, a data set with low-resolution images and high levels of noise, in which GEA considerably outperforms existing NAS methods.

**Table 3**
Performance comparison of different NAS methods on TransNAS-Bench-101. The first block shows the results for directly searching on each task. The second block shows the transferred versions of different methods, which are pretrained on the least time-consuming task, i.e., Jigsaw. The final row shows the possible best result in each task. Table adapted from [52, 53].

| | Tasks | Cls. Object | Cls. Scene | Autoencoding | Surf. Normal | Sem. Segment. | Room Layout | Jigsaw |
|---|---|---|---|---|---|---|---|---|
| | Metric | Acc. (%) ↑ | Acc. (%) ↑ | SSIM ↑ | SSIM ↑ | mIoU ↑ | L2 loss ↓ | Acc. (%) ↑ |
| Direct Search | RS [54] | 45.16±0.4 | 54.41±0.3 | 55.94±0.8 | 56.85±0.6 | 25.21±0.4 | 61.48±0.8 | 94.47±0.3 |
| | REA [44] | 45.39±0.2 | 54.62±0.2 | 56.96±0.1 | 57.22±0.3 | 25.52±0.3 | 61.75±0.8 | 94.62±0.3 |
| | PPO [55] | 45.19±0.3 | 54.37±0.2 | 55.83±0.7 | 56.90±0.6 | 25.24±0.3 | 61.38±0.7 | 94.46±0.3 |
| | DT | 42.03±5.0 | 49.80±8.6 | 51.20±3.3 | 55.03±2.7 | 22.45±3.2 | 66.98±2.3 | 88.95±9.1 |
| | BONAS [56]† | 45.50 | 54.56 | 56.73 | 57.46 | 25.32 | 61.10 | 94.81 |
| | weakNAS [57]† | 45.66 | 54.72 | 56.77 | 57.21 | 25.90 | 60.31 | 94.63 |
| | Arch-Graph-single [53]† | 45.48 | 54.70 | 56.52 | 57.53 | 25.71 | 61.05 | 94.66 |
| | GEA (Ours) | 45.98±0.2 | 54.85±0.1 | 57.11±0.3 | 58.33±1.0 | 25.95±0.2 | 59.93±0.5 | 94.96±0.2 |
| | GEA-Best (Ours)† | **46.32** | **54.94** | **57.72** | **59.62** | **26.27** | **59.38** | **95.37** |
| Transfer Search | REA-t [44] | 45.51±0.3 | 54.61±0.2 | 56.52±0.6 | 57.20±0.7 | 25.46±0.4 | 61.04±1.0 | - |
| | PPO-t [55] | 44.81±0.6 | 54.15±0.5 | 55.70±1.5 | 56.60±0.7 | 24.89±0.5 | 62.01±1.0 | - |
| | CATCH [58] | 45.27±0.5 | 54.38±0.2 | 56.13±0.7 | 56.99±0.6 | 25.38±0.4 | 60.70±0.7 | - |
| | BONAS-t [56]† | 45.38 | 54.57 | 56.18 | 57.24 | 25.24 | 60.93 | - |
| | weakNAS-t [57]† | 45.29 | 54.78 | 56.90 | 57.19 | 25.41 | 60.70 | - |
| | Arch-Graph-zero [53]† | 45.64 | 54.80 | 56.61 | 57.90 | 25.73 | 60.21 | - |
| | Arch-Graph [53]† | 45.81 | 54.90 | 56.58 | 58.27 | 25.69 | 60.08 | - |
| | GEA-t (Ours) | 46.03±0.3 | 54.86±0.1 | 56.99±0.3 | 58.21±0.9 | 25.88±0.3 | 59.85±0.5 | - |
| | GEA-t-Best (Ours)† | **46.32** | **54.94** | **57.72** | **59.62** | **26.27** | **59.38** | - |
| | Global Best | 46.32 | 54.94 | 57.72 | 59.62 | 26.27 | 59.38 | 95.37 |

† Results provided for the best run only.

Finally, we evaluate GEA on all 7 tasks from TransNAS-bench-101. Evaluating GEA on several tasks contributes to validating its generability and transferability across different problems, which is where NAS methods commonly fail [31, 30, 59]. For this, we conducted two different experiments: i) directly searching on each task independently; and ii) perform transfer search. For the latter, we followed common procedures [52], where we first search on jigsaw and use the final population as initialisation for the evolution when searching on the other tasks. The results for both experiments are shown in Table 3. From the results shown, it is possible to see that: first, directly searching on each task is an effective approach, where GEA is capable of achieving an higher mean performance, of 25 runs, higher than any other NAS method in all tasks. Also, when looking only at the best result, GEA is capable of achieving the best possible results in TransNAS-Bench-101, which means that GEA is capable of generating the most optimal architecture for each task. The same behaviours are present when transferring the knowledge from jigsaw to other tasks, where GEA-t achieves state-of-the-art results on all tasks. When compared to directly searching on each task, GEA-t has a slight improvement only on classification tasks, meaning that GEA does not require prior information to achieve state-of-the-art performances when compared to other NAS methods.

The obtained results in all the 3 benchmarks, which account to 11 different data sets, show that an evolutionary strategy, coupled with a mechanism to quickly evaluate architectures to guide the search, can achieve state-of-the-art results while still having competitive search times. Despite the complexity of search spaces and severe difficulty in obtaining their global information, the results show that guiding mechanisms powered by scoring architectures at initialisation stages have the advantage of acquiring preliminary information regarding in which direction the search should evolve. Therefore, GEA can quickly converge to better results by avoiding local minima, while still being efficient in terms of the required time.

### 4.3. Ablation Studies

This section extends the study about the importance of different parameters on GEA. The results, in terms of mean validation accuracy and standard deviation in NAS-Bench-201 CIFAR-10, can be seen in Table 4.

First, we look into the importance of the parameter $S$ by incrementally increasing its value from 1 to 10, and instead of randomly sampling $S$ architectures, replacing the pool of candidates, by simply selecting architectures from the population pool with highest and lowest fitnesses. For this, $P$ and $C$ were fixed to 10 and 200 respectively. From Table 4, a clear pattern can be seen, where the best results are obtained when $S$ is higher. Logically, sampling the lowest scoring individual to be the parent of the next generation yields the worst results, as this forces the evolution to follow the worst-known settings. $P = 10$ achieved a better mean validation accuracy than sampling the highest scoring individual. We justify that this is due to the fact that by having a high enough $P$ value, it allows that most of the time, one of the best architectures is chosen to be parent, while at the same time, it promotes exploration of the search space by not using the best-known setting yet. A visual representation of the evolution of the best architecture for the different parameter values, over 5 different runs, can be seen in Figure 5.

**Table 4**
Ablation studies for the number of parent candidates, $S$, the population size, $P$, and the regularization mechanism to remove individuals from the population. Results are shown in mean validation accuracy (%) and standard deviation from 5 runs in NAS-Bench-201 CIFAR-10 data set.

| Parameter | Value | Mean Validation Accuracy (%) |
|---|---|---|
| S | 1 | 91.09±0.45 |
|   | 3 | 91.45±0.20 |
|   | 5 | 91.41±0.24 |
|   | 7 | 91.47±0.16 |
|   | 10 | **91.56±0.05** |
|   | Highest | 91.50±0.19 |
|   | Lowest | 89.93±0.55 |
| P | 1 | 91.19±0.10 |
|   | 3 | 91.45±0.09 |
|   | 5 | **91.58±0.02** |
|   | 7 | 91.55±0.06 |
|   | 10 | 91.41±0.24 |
| Regularization | Oldest | **91.56±0.05** |
|   | Highest | 90.59±0.46 |
|   | Lowest | 91.30±0.23 |

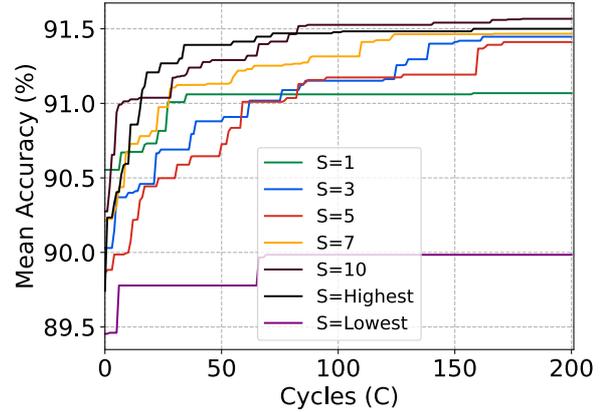

**Figure 5:** Mean validation accuracy (%) throughout the evolution for different parent sampling, $S$, schemes using NAS-Bench-201 CIFAR-10 for 5 runs.

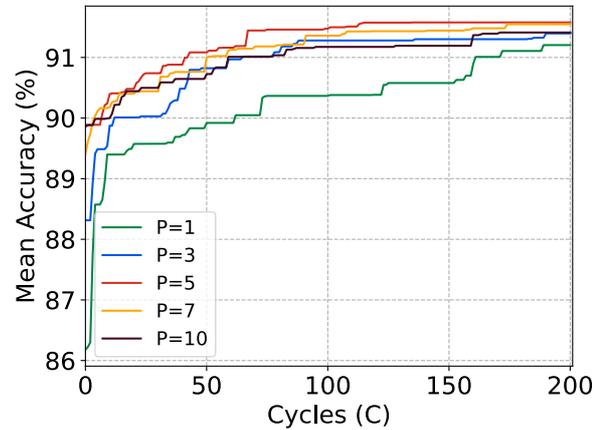

**Figure 6:** Mean validation accuracy (%) throughout the evolution for different population sizes, $P$, using NAS-Bench-201 CIFAR-10 for 5 runs.

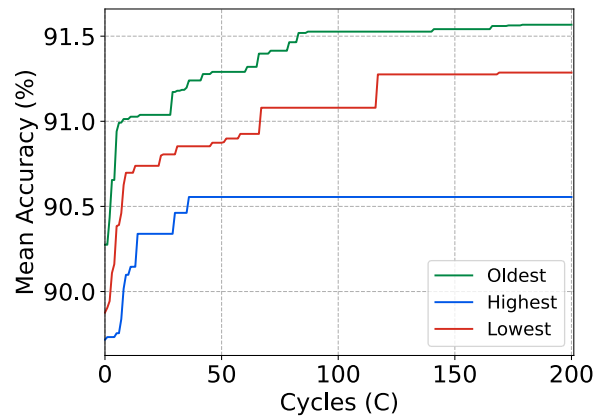

**Figure 7:** Mean validation accuracy (%) throughout the evolution for different regularization schemes using NAS-Bench-201 CIFAR-10 for 5 runs.

We also evaluate the importance of the population size, $P$. Similarly to $S$, we incrementally increase $P$ from 1 to 10. From Table 4, it is possible to see that higher values of $P$ achieve better results than lower values, and the best results are obtained with $P = 5$. This is because a smaller population size promotes exploitation, as the candidates sampled to be parents are more often among the best individuals, thus leading the search to better regions of the search space. In Figure 6 it is possible to see the evolution of the best architecture found by GEA for each $P$ value evaluated over 5 different runs.

Finally, we evaluated the importance of the regularization mechanism that removes individuals from the population. For this, we assessed the already discussed elimination by age by removing the oldest individual in the population and evaluated the results if the best (highest fitness) and the worst (lowest fitness) individuals were removed. From Table 4, it is clear that removing the best individual is the worst possible strategy, as it forces the search to ignore the best-known settings so far. Then, comparing removing the worst and the oldest, the best results are yielded when the oldest individual is removed, as it promotes further exploration of the search space. The results shown in Figure 7 for the evolution of the different regularization mechanisms clearly show that removing the oldest individuals yields the best results.

## 5. Conclusions

This paper proposes GEA, a guided evolution strategy for neural architecture search by leveraging zero-proxy estimation of untrained architectures. GEA forces exploitation of the most performant architectures by descendant generation and an exploration of the search space by conducting mutations. GEA guides the evolution by generating several architectures in each generation and evaluating them at the initialisation stage using a zero-proxy estimator, where only the highest-scoring architecture is trained and kept for the next generation. The generation of multiple architectures from an existing one in the population at each generation allows GEA to constantly extract knowledge concerning the search space without compromising the search itself, resulting in state-of-the-art performances in all data sets of NAS-Bench-101, NAS-Bench-201 and TransNAS-Bench-101 benchmarks.

The proposed guided NAS approach can be extended to multiple strategies, where the search method can be further improved by incorporating new regularisation and mutation mechanisms. Also, the components of the guiding mechanism can easily be transferred to other evolutionary algorithms, allowing existing NAS evolutionary methods to be further improved.

## Data availability

The data used in the paper is publically available datasets.

## Declaration of Competing Interest

The authors declare that they have no known competing financial interests or personal relationships that could have appeared to influence the work reported in this paper.

## Acknowledgments

This work was supported by 'FCT - Fundação para a Ciência e Tecnologia' through the research grants '2020.04588.BD' and 'UI/BD/150765/2020', partially supported by NOVA LINCS (UIDB/04516/2020) with the financial support of FCT, through national funds and CENTRO-01-0247-FEDER-113023 - DeepNeuronic.

## References

[1] L. Deng, D. Yu, et al., Deep learning: methods and applications, Foundations and Trends in Signal Processing (2014).
[2] I. Goodfellow, Y. Bengio, A. Courville, Y. Bengio, Deep learning, volume 1, MIT press Cambridge, 2016.
[3] A. Garcia-Garcia, S. Orts-Escolano, S. Oprea, V. Villena-Martinez, P. Martinez-Gonzalez, J. Garcia-Rodriguez, A survey on deep learning techniques for image and video semantic segmentation, Applied Soft Computing 70 (2018) 41–65.
[4] A. Conneau, H. Schwenk, L. Barrault, Y. LeCun, Very deep convolutional networks for text classification, in: M. Lapata, P. Blunsom, A. Koller (Eds.), EACL, 2017.
[5] A. Khan, A. Sohail, U. Zahoora, A. S. Qureshi, A survey of the recent architectures of deep convolutional neural networks, Artificial Intelligence Review (2020).
[6] A. Krizhevsky, I. Sutskever, G. E. Hinton, Imagenet classification with deep convolutional neural networks, in: Advances in Neural Information Processing Systems, 2012.
[7] C. Szegedy, W. Liu, Y. Jia, P. Sermanet, S. E. Reed, D. Anguelov, D. Erhan, V. Vanhoucke, A. Rabinovich, Going deeper with convolutions, in: CVPR, 2015.
[8] K. He, X. Zhang, S. Ren, J. Sun, Deep residual learning for image recognition, in: CVPR, 2016.
[9] G. Huang, Z. Liu, L. van der Maaten, K. Q. Weinberger, Densely connected convolutional networks, in: CVPR, 2017.
[10] F. Chollet, Xception: Deep learning with depthwise separable convolutions, in: CVPR, 2017, pp. 1251–1258.
[11] M. Tan, Q. V. Le, Efficientnet: Rethinking model scaling for convolutional neural networks, in: K. Chaudhuri, R. Salakhutdinov (Eds.), ICML, 2019.
[12] A. Dosovitskiy, L. Beyer, A. Kolesnikov, D. Weissenborn, X. Zhai, T. Unterthiner, M. Dehghani, M. Minderer, G. Heigold, S. Gelly, J. Uszkoreit, N. Houlsby, An image is worth 16x16 words: Transformers for image recognition at scale, in: ICLR, 2021.
[13] T. Elsken, J. H. Metzen, F. Hutter, Neural architecture search: A survey, Journal of Machine Learning Research (2019).
[14] M. Wistuba, A. Rawat, T. Pedapati, A survey on neural architecture search, CoRR abs/1905.01392 (2019).
[15] V. Lopes, L. A. Alexandre, Auto-classifier: A robust defect detector based on an automl head, in: ICONIP, Springer, Cham, 2020, pp. 137–149.
[16] V. Lopes, A. Gaspar, L. A. Alexandre, J. Cordeiro, An automl-based approach to multimodal image sentiment analysis, in: IJCNN, 2021.
[17] R. Ru, P. M. Esperança, F. M. Carlucci, Neural architecture generator optimization, in: H. Larochelle, M. Ranzato, R. Hadsell, M. Balcan, H. Lin (Eds.), NeurIPS, 2020.
[18] V. Lopes, L. A. Alexandre, Towards less constrained macro-neural architecture search, arXiv preprint arXiv:2203.05508 (2022).
[19] D. Baymurzina, E. A. Golikov, M. S. Burtsev, A review of neural architecture search, Neurocomputing 474 (2022) 82–93.
[20] B. Zoph, Q. V. Le, Neural architecture search with reinforcement learning, in: ICLR, 2017.
[21] C. Liu, B. Zoph, M. Neumann, J. Shlens, W. Hua, L.-J. Li, L. Fei-Fei, A. Yuille, J. Huang, K. Murphy, Progressive neural architecture search, in: ECCV, 2018.
[22] B. Zoph, V. Vasudevan, J. Shlens, Q. V. Le, Learning transferable architectures for scalable image recognition, CVPR (2018).
[23] Y. Shu, W. Wang, S. Cai, Understanding architectures learnt by cell-based neural architecture search, in: 8th International Conference on Learning Representations, ICLR, OpenReview.net, 2020.
[24] F. M. Carlucci, P. M. Esperança, M. Singh, V. Gabillon, A. Yang, H. Xu, Z. Chen, J. Wang, Manas: Multi-agent neural architecture search, arXiv preprint arXiv:1909.01051 (2019).
[25] K. Jing, J. Xu, Z. Zhang, A neural architecture generator for efficient search space, Neurocomputing 486 (2022) 189–199.
[26] C. White, A. Zela, R. Ru, Y. Liu, F. Hutter, How powerful are performance predictors in neural architecture search?, in: NeurIPS, 2021.

[27] H. Liu, K. Simonyan, Y. Yang, DARTS: Differentiable Architecture Search, in: ICLR, 2019.
[28] H. Pham, M. Guan, B. Zoph, Q. Le, J. Dean, Efficient neural architecture search via parameters sharing, in: ICML, 2018.
[29] Y. Xu, L. Xie, X. Zhang, X. Chen, G. Qi, Q. Tian, H. Xiong, PC-DARTS: partial channel connections for memory-efficient architecture search, in: ICLR, 2020.
[30] X. Dong, Y. Yang, NAS-Bench-201: Extending the Scope of Reproducible Neural Architecture Search, in: ICLR, 2020.
[31] A. Yang, P. M. Esperança, F. M. Carlucci, Nas evaluation is frustratingly hard, in: ICLR, 2020.
[32] V. Lopes, S. Alirezazadeh, L. A. Alexandre, EPE-NAS: Efficient Performance Estimation Without Training for Neural Architecture Search, in: ICANN, 2021.
[33] J. Mellor, J. Turner, A. J. Storkey, E. J. Crowley, Neural Architecture Search without Training, in: ICML, 2021.
[34] X. Ning, C. Tang, W. Li, Z. Zhou, S. Liang, H. Yang, Y. Wang, Evaluating efficient performance estimators of neural architectures, in: NeurIPS, 2021.
[35] Z. Zhong, J. Yan, W. Wu, J. Shao, C.-L. Liu, Practical block-wise neural network architecture generation, in: CVPR, 2018.
[36] Y. Li, M. Dong, Y. Xu, Y. Wang, C. Xu, Neural architecture tuning with policy adaptation, Neurocomputing 485 (2022) 196–204.
[37] X. Dong, Y. Yang, Searching for a robust neural architecture in four gpu hours, in: CVPR, 2019, pp. 1761–1770.
[38] L. Li, A. Talwalkar, Random search and reproducibility for neural architecture search, in: UAI, PMLR, 2020, pp. 367–377.
[39] X. Dong, Y. Yang, One-Shot Neural Architecture Search via Self-Evaluated Template Network, in: ICCV, IEEE, 2019.
[40] A. Zela, T. Elsken, T. Saikia, Y. Marrakchi, T. Brox, F. Hutter, Understanding and robustifying differentiable architecture search, in: ICLR, 2020.
[41] K. O. Stanley, R. Miikkulainen, Evolving neural networks through augmenting topologies, Evolutionary computation 10 (2002) 99–127.
[42] E. Real, S. Moore, A. Selle, S. Saxena, Y. L. Suematsu, J. Tan, Q. V. Le, A. Kurakin, Large-scale evolution of image classifiers, in: D. Precup, Y. W. Teh (Eds.), ICML, 2017.
[43] T. Elsken, J. H. Metzen, F. Hutter, Efficient multi-objective neural architecture search via lamarckian evolution, in: ICLR, 2019.
[44] E. Real, A. Aggarwal, Y. Huang, Q. V. Le, Regularized Evolution for Image Classifier Architecture Search, in: AAAI, 2019, pp. 4780–4789.
[45] H. Liu, K. Simonyan, O. Vinyals, C. Fernando, K. Kavukcuoglu, Hierarchical Representations for Efficient Architecture Search, in: ICLR, 2018.
[46] Y. Liu, Y. Sun, B. Xue, M. Zhang, G. G. Yen, K. C. Tan, A survey on evolutionary neural architecture search, IEEE Trans. Neural Networks Learn. Syst. (2021).
[47] X. Xie, Y. Liu, Y. Sun, G. G. Yen, B. Xue, M. Zhang, Benchenas: A benchmarking platform for evolutionary neural architecture search, arXiv:2108.03856 (2021).
[48] C. Wei, C. Niu, Y. Tang, J. Liang, NPENAS: neural predictor guided evolution for neural architecture search, CoRR abs/2003.12857 (2020).
[49] P. Bashivan, M. Tensen, J. J. DiCarlo, Teacher guided architecture search, in: ICCV, 2019.
[50] K. Yu, R. Ranftl, M. Salzmann, Landmark regularization: Ranking guided super-net training in neural architecture search, in: CVPR, 2021.
[51] C. Ying, A. Klein, E. Christiansen, E. Real, K. Murphy, F. Hutter, NAS-Bench-101: Towards Reproducible Neural Architecture Search, in: ICML, 2019.
[52] Y. Duan, X. Chen, H. Xu, Z. Chen, X. Liang, T. Zhang, Z. Li, Transnas-bench-101: Improving transferability and generalizability of cross-task neural architecture search, in: CVPR, 2021.
[53] M. Huang, Z. Huang, C. Li, X. Chen, H. Xu, Z. Li, X. Liang, Archgraph: Acyclic architecture relation predictor for task-transferable neural architecture search, CoRR abs/2204.05941 (2022).
[54] J. Bergstra, Y. Bengio, Random search for hyper-parameter optimization, J. Mach. Learn. Res. 13 (2012) 281–305.
[55] J. Schulman, F. Wolski, P. Dhariwal, A. Radford, O. Klimov, Proximal policy optimization algorithms, CoRR abs/1707.06347 (2017).
[56] H. Shi, R. Pi, H. Xu, Z. Li, J. T. Kwok, T. Zhang, Bridging the gap between sample-based and one-shot neural architecture search with BONAS, in: H. Larochelle, M. Ranzato, R. Hadsell, M. Balcan, H. Lin (Eds.), NeurIPS, 2020.
[57] J. Wu, X. Dai, D. Chen, Y. Chen, M. Liu, Y. Yu, Z. Wang, Z. Liu, M. Chen, L. Yuan, Stronger NAS with weaker predictors, in: NeurIPS, 2021.
[58] X. Chen, Y. Duan, Z. Chen, H. Xu, Z. Chen, X. Liang, T. Zhang, Z. Li, CATCH: context-based meta reinforcement learning for transferrable architecture search, in: ECCV, 2020.
[59] X. Wan, B. Ru, P. M. Esperança, Z. Li, On redundancy and diversity in cell-based neural architecture search, in: ICLR, 2022.